# Convolutional Neural Network Pruning Using Filter Attenuation

*Morteza Mousa-Pasandi[1], Mohsen Hajabdollahi[1], Nader Karimi[1], Shadrokh Samavi[1,2], Shahram Shirani[2]*

[1]Department of Electrical and Computer Engineering, Isfahan University of Technology, Isfahan, 84156-83111 Iran
[2]Department of Electrical and Computer Engineering, McMaster University, Hamilton, ON L8S 4L8, Canada

**ABSTRACT**

Filters are the essential elements in convolutional neural networks (CNNs). Filters are corresponded to the feature maps and form the main part of the computational and memory requirement for the CNN processing. In filter pruning methods, a filter with all of its components, including channels and connections, are removed. The removal of a filter can cause a drastic change in the network's performance. Also, the removed filters can't come back to the network structure. We want to address these problems in this paper. We propose a CNN pruning method based on filter attenuation in which weak filters are not directly removed. Instead, weak filters are attenuated and gradually removed. In the proposed attenuation approach, weak filters are not abruptly removed, and there is a chance for these filters to return to the network. The filter attenuation method is assessed using the VGG model for the Cifar10 image classification task. Simulation results show that the filter attenuation works with different pruning criteria, and better results are obtained in comparison with the conventional pruning methods.

## 1. INTRODUCTION

Convolutional neural networks (CNNs) are structures with strong feature extraction capabilities that are widely used in pattern recognition and object classification [1-2]. There is a large amount of computational complexity due to the filtering implantation and many memory operations due to feature maps storing [3]. For example, in VGG16 and ResNet-18, 15.2M and 14.2M parameters are employed, respectively. Implementing such a massive model on devices with limited computational resources can be a problem [4].

Several research areas are emerging to deal with the mentioned above problems, including quantization, factorization, and pruning [5]. Quantization methods employ various techniques to lower the number of bits required for representing the CNN parameters and reduce the total computation bandwidth [6][7][8]. The factorization methods are utilized with different encoding techniques to obtain a reduced model without loss of accuracy [9] [10]. It is confirmed that the CNN structures have many redundant parameters [11], [12]. Pruning in which un-necessary parameters are removed from the network model has been attracted a lot of researches since 1990 [13][14]. Recently the pruning techniques are widely adopted for CNN structures. By pruning, parameters, computations, and memory consumption could be reduced. Furthermore, pruning has the advantages of increasing the speed of the training and the inference phase in GPU and CPU [15][16]. CNN structures consist of different abstraction levels. Therefore, different levels can be targeted in pruning, including connection, channel, and filter.

Connection pruning emphasizes on the pruning of the weak connections, which are not useful for the network structure [17][18]. To have a better sparsity suitable for efficient implementation, channel pruning is introduced and investigated by different researchers [15], [19], [20]. In the channel pruning methods, weak channels are determined by all of their connections and all of them are removed by pruning. Although in the channel pruning, better sparsity is provided, removal of a channel necessarily does not result in feature map pruning [20]. To address this problem, filter pruning is proposed, which corresponds to the feature map pruning. However, filters are the main elements of CNNs, and their pruning can lead to considerable variations in the network results. Also, filters selected for pruning are unimportant in the current structure, and their importance may change during the next pruning steps. Therefore, the removed filters do not have any chance to come back to the network structure.

In this paper, the problems mentioned above in the filter pruning are addressed. An attenuation approach for filter pruning is proposed in which the weak filters instead of complete removal are attenuated. Attenuation has two advantages. First, the effect of the filter to be pruned is reduced by attenuation; second, an attenuated filter can recover from attenuation in the next round of pruning and becomes an important filter. This algorithm is appropriate for those elements of CNN, which pruning them causes a harsh variation to the model's performance.

The remaining part of this paper is organized as follows. In Section 2, filter pruning is briefly presented. In Section 3, the proposed method for filter attenuation is explained in detail. In Section 4, filter attenuation is formulated to provide a better illustration. In Section 5, the proposed method is assessed via a practical experiment, and corresponding results are reported. Finally, in Section 6, the conclusion is presented.

## 2. FILTER PRUNING

Feature maps can be regarded as the most memory consuming elements in CNN structures. In the filter pruning methods, filters with all of their channels and connections are removed, which leads to the removal of corresponding feature maps. In Fig. 1, one layer of a sample CNN with eight filters is illustrated. Filter pruning is conducted for number 2 and 7 filters, which are shown with red dotted lines. Removing a filter is accompanied by removing the corresponding feature maps, as illustrated in Fig. 1.

There are different methods studied for the problem of filter pruning [12], [21], [22]. Several criteria are introduced to select filters for pruning. L1 norm in [23] is used, and this criterion is improved by using the L2 norm in [24]. Also, in [25], it was stated that the cosine distance metric has better results. Filters could be pruned in different ways. In the filter pruning, better sparsity and structural regularity are provided, however by removing a filter some problems arose. The first one is accuracy drop due to the pruning because filters are the main element in the CNN structure [23]. Also, another factor in accuracy reduction is the effect of each filter on the next layer channels. The second one is the lack of a method giving the pruned filter a recovery possibility. In each pruning round, a new structure is created and pruned filters in the new structures may become significant. Filter pruning by itself and permanent removal of a filter can lead to an accuracy reduction problem. In the previous studies, the accuracy drop is compensated mainly by fine-tuning [23], [26].

## 3. PRUNING WITH FILTER ATTENUATION

In Fig. 2, the proposed method for CNN pruning based on filter attenuation is illustrated. The core of the proposed method is based on two ideas. The first one is gradually removing weak filters to alleviate the effect of their removal on the training performance. Second, granting a possibility of coming back to the network structure to those filters which are selected for pruning. To realize these ideas, an attenuation-based pruning is proposed. In the filter attenuation, filters that are selected to be pruned are not set to zero but they are attenuated by an attenuation coefficient. From one side, using attenuation the weak filters are punished and gradually tend to zero. On the other hand, attenuated filters are not excluded from the network structure and can recover their importance. In Fig. 2, on the left side, filter attenuation is illustrated. The more important filters and feature maps are colored darker. Filters and feature maps with red dotted lines indicate those that are selected for pruning. Suppose that two filters are selected for attenuation in each step. Filter numbers 7 and 2 are selected for pruning. As illustrated on the left side of Fig. 2, filters and therefore their corresponding feature maps selected for pruning are only attenuated. They are not removed, unlike the pruned filters illustrated in Fig. 1. After fine-tuning, filter importance can be changed, as shown on the right side of Fig. 2. As illustrated in the right side of Fig. 2, the importance of the filter with number 2, which was attenuated is changed and moved to the set of filters that are not attenuated. Instead, the filter with number 5 is replaced and selected for attenuation. The filter attenuation algorithm is illustrated in Fig. 3, as pseudo-code.

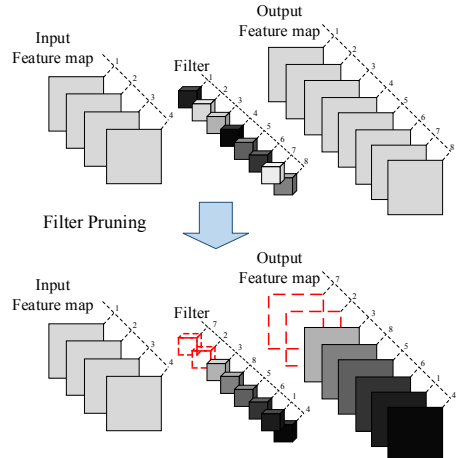

Fig. 1. Filter pruning, red dotted lines indicate pruned filters.

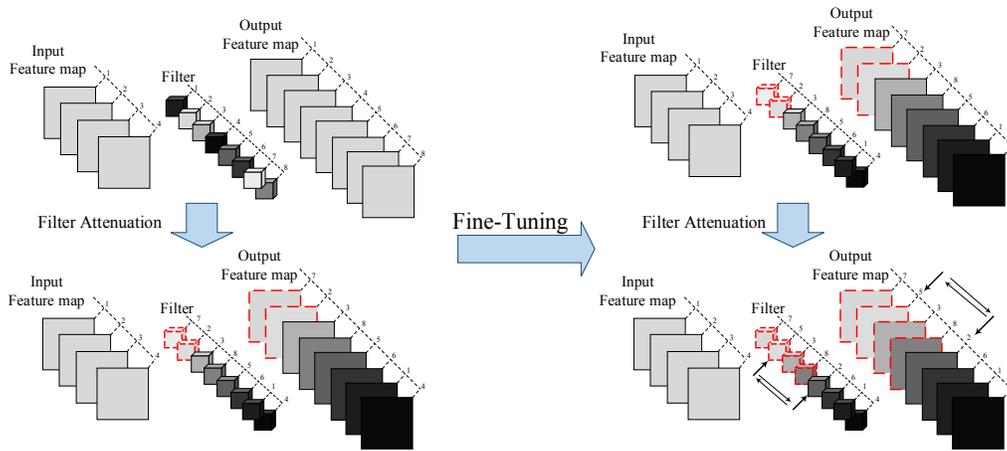

Fig. 2. Filter attenuation, more important filters are colored darker. Dotted red lines indicates elements which are attenuated. Left, filter attenuation which not lead to a filter removal, right: fine tuning which exchange an attenuated filter with another one.

As shown in Fig. 3, after training the network structure, attenuation is conducted by a parameter $k$, which is incremented by $a$. After attenuation, filters which have an importance e.g. L1-Norm, less than a threshold are removed. This threshold is set so small that means pruning is conducted for those filters, which are zero in practice. By establishing an appropriate threshold for the proposed algorithm, the number of filters that are attenuated is larger than the pruned filters in each step. In this way, a gradual pruning, with a possible recovery for the attenuated filters, becomes feasible. During each step of the algorithm, the weak filters are attenuated, and some previously attenuated filters can be attenuated or pruned. Also, an attenuated filter can be recovered and regain its status as an important filter.

## 4. FILTER ATTENUATION FORMULATION

Suppose that all of the filters in the $l$th layer are represented by $F^l \in \mathbb{R}^{N_l \times N_{l+1} \times K \times K}$, in which $N_l$ is the number of input feature maps in the $l$th layer, and k is the kernel size. An output feature map resulted from the convolutional operations can be computed using equation (1).

$$X_i^{l+1} = \sum_{j=1}^{N_i} X_j^l \otimes F_{i,j}^l \tag{1}$$

In which $X_i^l$ is the $i$th feature map in the $l$th layer, $F_{i,j}^l$ is the $j$th channel of the $i$th filter in the $l$th layer. Considering the $l$th layer of a CNN, filter pruning selects a filter set for pruning. Pruning a filter corresponds to the pruning of the corresponding feature maps in the next layer. Therefore, the set of pruned feature maps can be defined as a set $R = \{X_x^{l+1}, \dots, X_{|D|}^{l+1}\}$. There are $N_{l+1}$ feature maps in the $(l+1)$th layer, and $R$ represent a set of them with a cardinality |D|. Filter pruning can be interpreted as a procedure that aims to remove those feature maps which have the least effect on the producing the $(l+2)$th layer's feature map. It can be said that filters in the $l$th layer are removed, which are correspond to the $(l+1)$th layer, and the effect of this removal can be minimized in making the $(l+2)$th feature maps. Therefore, the selection of set R of filters is targeted which satisfies equation (2).

$$\min_{R \in F^l} \left( \sum_{i=1}^{N_{l+1}} \sum_{j \mid X_x^{l+1} \in R} X_j^{l+1} \otimes F_{i,j}^{l+1} \right) \tag{2}$$

Equation (2) represents that, the sum of variations in the feature maps in the $(l+2)$th layer should be minimized, which could mean filters with small values are suitable for pruning. In the pruning process, filters are selected based on different criteria, including L1-Norm, cosine distance, etc. But after pruning, filter values become zero as equation (3), in which $\alpha$ is the pruning threshold. Equation (3) is an example of filter pruning by value. Considering the pruning mask, filters in the $l$th layer are updated using equation (4).

$$Mask_i^l = \begin{cases} 0 & \sum_{j=1}^{N_i} F_{i,j}^l < \alpha * \frac{1}{N_i} \sum_{i=1}^{N_i} \sum_{j=1}^{N_{i+1}} F_{i,j}^l \\ 1 & O.W \end{cases} \tag{3}$$

$$F_{i,j}^l = F_{i,j}^l - \eta \left( \frac{\partial C}{\partial F_{i,j}^l} * Mask_i^l \right) \tag{4}$$

In equation (4), $\eta$ is learning rate, and $\frac{\partial C}{\partial F_{i,j}^l}$ is the error of the model with respect to the $F_{i,j}^l$. As illustrated in equation (3) filters that are selected for pruning becomes zero. Using filter attenuation, filters are attenuated using an attenuate factor $F_a$, but filters that have very small value becomes zero as equation (5).

$$Mask_i^l = \begin{cases} 0 & \sum_{j=1}^{N_i} F_{i,j}^l < \beta * \frac{1}{N_i} \sum_{i=1}^{N_i} \sum_{j=1}^{N_{i+1}} F_{i,j}^l \\ F_a & \sum_{j=1}^{N_i} F_{i,j}^l < \alpha * \frac{1}{N_i} \sum_{i=1}^{N_i} \sum_{j=1}^{N_{i+1}} F_{i,j}^l \\ 1 & Otherwise \end{cases} \tag{5}$$

where $\alpha$ and $\beta$ are attenuation and pruning thresholds, respectively, and filters are updated as equation (4). Using equation (5) under the proposed algorithm in Fig. 3, weak filters are gradually discarded from the training, and weakness or robustness of all filters can be modified.

---

Input: training data, base model architecture, permitted accuracy drop: $T_1$, pruning rate: $T_2$, attenuation parameter: $k$.
Output: Pruned model architecture

1: Train model for a few epochs
2: **While** all the filters are not pruned
3:     $k = k + a$
4:     Calculate the L1-Norm for each filter
5:     attenuate the least k important filters
6:     fine tuning for a few epochs
7:     Calculate the L1-Norm for each filter
8:     Prune filters with L1-Norm < $T_2$
9:     **If** accuracy drop < $T_1$
10:         fine tuning in a few epochs
11:     **If** accuracy is not acceptable
12:         Recover the last pruned filters
13:         Fine tuning in a few epochs
14:         Go to **End**
15:     **If** model complexity is acceptable
16:         Save model and go to **End**
17: **End while**
18: **End**

Fig. 3. Pseudo code of pruning using filter attenuation.

## 5. EXPERIMENTAL RESULTS

A VGG16 network to classify Cifar10 images is trained, and then the network structure is pruned by different pruning criteria. L1-Norm and using standard deviation (STD) of the filters are two effective ways for pruning, which were analyzed during our experiments. To evaluate the pruning performance, the amount of model reduction size, as well as the resulted accuracy, are reported. All of the experiments are conducted based on the Python programming language employing the TensorFlow framework. A computer that has an Nvidia GPU 1080 Ti is used as an accelerator for training and testing the experimented models. Attenuation factor $F_a$ and $a$ in pseudo-code illustrated in Fig. 3 is considered equal to 0.8, and 100, respectively

In Fig. 4 the trend of pruning by different criteria is illustrated. The illustrations in Fig. 4 are based on the number of pruned filters with respect to the resulted accuracy. In Fig. 4a, the results of the proposed pruning with the L1-norm criteria are illustrated. It can be observed that, with the same number of pruned filters, better accuracies are observed. Also, similar results are observed for the pruning by the STD criteria as well as the filter attenuation using STD criteria. Results of pruning illustrated in Fig. 4, indicate that filter attenuation, as a method of pruning, improves the result of previous pruning methods on different criteria.

In Table. 1, profiles of pruned filters, after 50% pruning, are illustrated. In some layers, the small number of filters are pruned, while in some of the layers more filters are pruned. The number of pruned filters in a layer can be used to identify the importance of that layer. By using the L1-Norm, the first layers are regarded as important while using STD, the last layers are considered as significant. Some differences are observed in the case of pruning by the proposed method, which is due to the recovery possibility of the attenuated filters.

Table 1. Profiles of pruned filters resulted from different methods

| Layer | Original | L1-Norm | Proposed + L1-Norm | STD | Proposed+ STD |
|---|---|---|---|---|---|
| Conv-1 | 64 | 4 | 7 | 54 | 53 |
| Conv-2 | 64 | 6 | 4 | 55 | 55 |
| Conv-3 | 128 | 24 | 33 | 117 | 119 |
| Conv-4 | 128 | 19 | 5 | 51 | 44 |
| Conv-5 | 256 | 30 | 83 | 105 | 103 |
| Conv-6 | 256 | 367 | 81 | 93 | 99 |
| Conv-7 | 256 | 31 | 78 | 97 | 94 |
| Conv-8 | 512 | 115 | 160 | 118 | 116 |
| Conv-9 | 512 | 350 | 330 | 325 | 321 |
| Conv-10 | 512 | 382 | 353 | 464 | 446 |
| Conv-11 | 512 | 392 | 335 | 359 | 376 |
| Conv-12 | 512 | 372 | 319 | 94 | 106 |
| Conv-13 | 512 | 350 | 324 | 180 | 180 |

After 42% filter pruning (pruning the 1800 filters in the VGG model), there are some filters that were attenuated at least one time but currently are remaining. In Fig. 5, the number of filters that are remained but previously were attenuated in term of the number of attenuations are illustrated. For example, in Fig. 5, 10 filters were attenuated four times. This experiment demonstrates that a number of filters can be recovered from the pruning. These filters currently are important while previously were identified as unimportant.

## 6. CONCLUSION

A pruning method based on filter attenuation was proposed. During pruning, filters selected for pruning were attenuated instead of abruptly being excluded from the network structure. In this method, two key features can be preserved. First, filters are pruned gradually. Second, among a pool of important and non-important filters, those that are selected as non-important have a possibility to be important. The simulation results showed that the proposed pruning method works on L1-norm and STD based filter pruning and improves their result about %0.2 on average. Also, simulation results demonstrated that a lot of attenuated filters were finally considered as important filters.

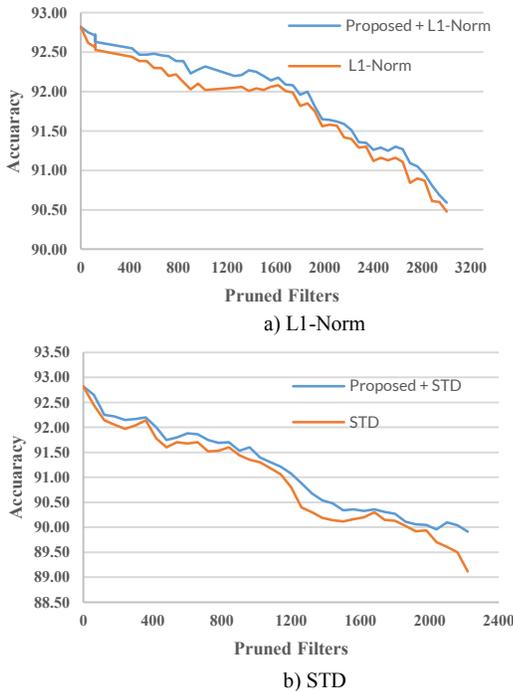

a) L1-Norm

b) STD

Fig. 4. Results of the proposed method with different criteria

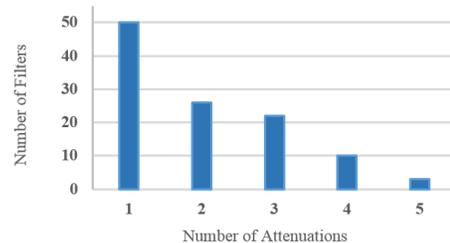

Fig. 5. The number of filters that are attenuated in terms of the number of attenuations.